\documentclass[twoside,11pt]{article}

\usepackage[preprint]{jmlr2e_preprint}

\usepackage{lastpage}
\jmlrheading{XX}{2025}{1--\pageref{LastPage}}{04/25; Revised XX/XX}{X/XX}{XX-XXXX}{Dempster et al}

\usepackage[utf8]{inputenc}
\usepackage[T1]{fontenc}
\usepackage{hyperref}
\usepackage{url}
\usepackage{booktabs}
\usepackage{nicefrac}
\usepackage{microtype}
\usepackage{xcolor}

\usepackage{graphicx}
\usepackage{mathtools}

\usepackage{algorithm}
\usepackage{algorithmic}

\usepackage{bm}

\DeclareMathOperator*{\argmin}{arg\,min}
\DeclareMathOperator*{\argmax}{arg\,max}

\newcommand{\bX}{{\bf X}}

\newcommand{\by}{{\bf y}}
\newcommand{\bH}{{\bf H}}
\newcommand{\bx}{{\bf x}}
\newcommand{\bI}{{\bf I}}
\newcommand{\bU}{{\bf U}}
\newcommand{\bV}{{\bf V}}
\newcommand{\bs}{{\bf s}}
\newcommand{\bZ}{{\bf Z}}
\newcommand{\bz}{{\bf z}}
\newcommand{\bA}{{\bf A}}
\newcommand{\bW}{{\bf W}}

\newcommand{\bv}{{\bf v}}

\newcommand{\bSigma}{\bm{\Sigma}}
\newcommand{\bbeta}{\bm{\beta}}
\newcommand{\balpha}{\bm{\alpha}}
\newcommand{\hbbeta}{\hat{\bm{\beta}}}
\newcommand{\tbbeta}{\tilde{\bm{\beta}}}

\newcommand{\T}{{\rm T}}

\newif\ifexclude

\newcommand{\pv}{\textsc{PreVal}}
\newcommand{\lr}{LR}

\newcommand{\zo}{\mbox{0--\!1}~loss}

\newcommand{\minirocket}{\textsc{MiniRocket}}

\ShortHeadings{PreVal}{Dempster et al}
\firstpageno{1}

\begin{document}

\title{Prevalidated Ridge Regression is a Highly-Efficient Drop-In Replacement for Logistic Regression for High-Dimensional Data}

\author{%
\name Angus Dempster \email angus.dempster@monash.edu \\%
\name Geoffrey I. Webb \\
\name Daniel F. Schmidt \\
\addr Monash University, Melbourne
}

\editor{TBA}

\maketitle

\begin{abstract}%
    Logistic regression is a ubiquitous method for probabilistic classification.  However, the effectiveness of logistic regression depends upon careful and relatively computationally expensive tuning, especially for the regularisation hyperparameter, and especially in the context of high-dimensional data.  We present a prevalidated ridge regression model that closely matches logistic regression in terms of {\zo} and log-loss, particularly for high-dimensional data, while being significantly more computationally efficient and having no user-tuned hyperparameters (the regularisation hyperparameter is learned automatically as part of the fitting process).  We scale the coefficients of the model so as to minimise log-loss for a set of prevalidated predictions derived from the estimated leave-one-out cross-validation error.  This exploits quantities already computed in the course of fitting the ridge regression model in order to find the scaling parameter with nominal additional computation.%
\end{abstract}%

\begin{keywords}
    probabilistic classification, prevalidation, ridge regression, high-dimensional data, leave-one-out cross-validation
\end{keywords}


\section{Introduction} \label{sec-introduction}

Ridge regression \citep{hoerl_and_kennard_1970} can be used as a classifier by encoding the class as a regression target \citep[e.g.][pp 184--186]{bishop_2006}, and has an efficient closed-form solution that allows for evaluating multiple candidate values of the ridge parameter with a single model fit \citep[e.g.,][]{tew_etal_2023}.  However, by itself, ridge regression does not provide probabilistic predictions.

In contrast, fitting a logistic regression model involves directly optimising log-loss.  However, regularised maximum likelihood estimates for logistic regression can only be found by an iterative solver.  Additionally, and in contrast to ridge regression, evaluating different candidate values for an $\ell_{q}$ regularisation parameter requires refitting the model, possibly multiple times for each candidate value of the regularisation parameter (e.g., under $k$-fold cross-validation).

We show that a ridge regression model, with its coefficients scaled so as to minimise log-loss for a set of prevalidated predictions---derived from the estimated leave-one-out cross-validation (LOOCV) error---closely matches logistic regression in terms of both {\zo} (classification error) and log-loss, while being significantly more computationally efficient.

This approach has several distinct advantages.  It is highly efficient, making use of quantities already computed in fitting the ridge regression model in order to find the scaling parameter, $\kappa$, and tune the probabilities of the model.  The model fit is deterministic (for a given training set), as it is based on LOOCV, without the variability inherent in $k$-fold ($k < n$) cross-validation.  Significantly, the process also has essentially no hyperparameters: while the set of candidate values for the ridge parameter is technically a hyperparameter, in practice this can be kept at its default value.

While the shortcut method for computing LOOCV error for ridge regression and probability calibration are both individually well established, prior work does not appear to take advantage of the shortcut estimate of LOOCV error for the purposes of calibrating a ridge regression model, or to recognise how broadly effective the resulting model is.  The potential advantages of the prevalidated ridge regression model over logistic regression in terms of computational efficiency, simplicity, and reliability are significant, because logistic regression is so widely used, and because logistic regression is sensitive to the correct tuning of its hyperparameters, particularly in relation to regularisation, and particularly in the context of high-dimensional data.

We demonstrate empirically, with results for 273 datasets drawn from multiple domains, that the resulting prevalidated ridge regression model ({\pv}) is highly effective as a drop-in replacement for regularised maximum likelihood logistic regression ({\lr}), particularly for high-dimensional data, while being, on average, multiple orders of magnitude more computationally efficient (up to approximately $1{,}000\times$ faster to train compared to an off-the-shelf implementation of logistic regression, depending on the dataset).

\begin{figure}%
    \centering%
    \includegraphics[width=\linewidth]{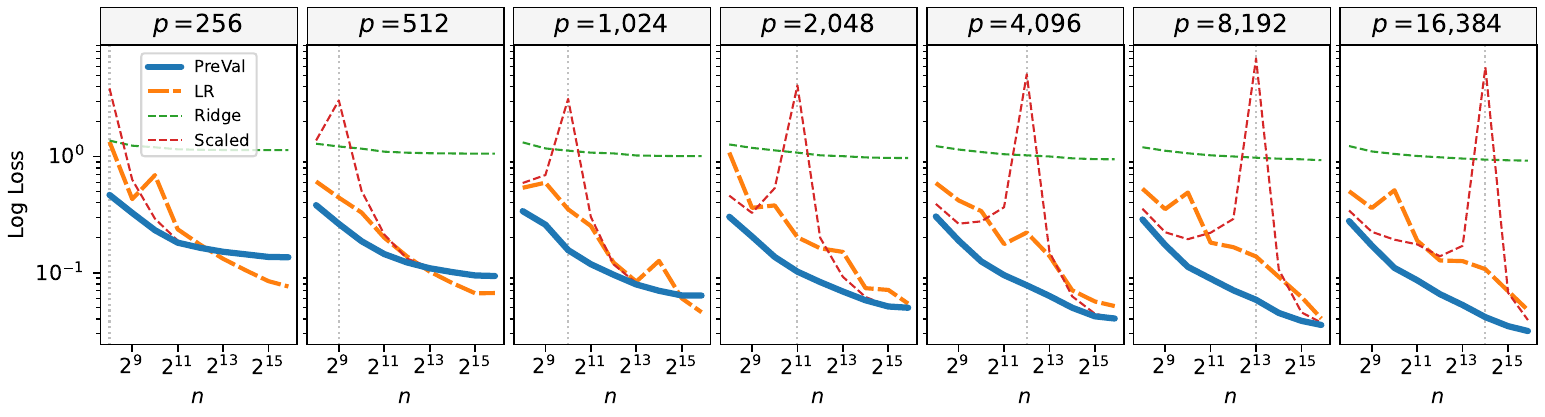}%
    \vspace{-1em}%
    \caption{Learning curves (log-loss) for {\pv} (blue), {\lr} (orange), ridge regression (green), and na\"{i}vely scaled ridge regression (red), for increasing numbers of features, $p \in \{2^{8}, 2^{9}, \dots, 2^{14}\}$, for a random projection of the MNIST dataset.  {\pv} requires a fraction of the compute while closely matching the log-loss of {\lr} in most scenarios, often showing advantage when $p$ is large relative to $n$.}%
    \label{fig-mnist-projection-loss}%
\end{figure}

Figure \ref{fig-mnist-projection-loss} shows learning curves (log-loss) for {\pv} (blue) versus {\lr} (orange), as well as for a `na\"{i}vely' scaled version of ridge regression (red) and a standard ridge regression model (green), on a random projection of the MNIST dataset.  Log-loss for {\pv} approaches and then betters that of {\lr} as the number of features grows.  (Inconsistency in the log-loss curves for {\lr} appears to be due to overfitting for certain values of $p$ and $n$, possibly exacerbated by variability from internal cross-validation.)  In fact, {\pv} closely matches both the \mbox{0--\!1} and log-loss of {\lr} on a large number of datasets: see Section \ref{sec-experiments}.  Without scaling, ridge regression yields poor (high) log-loss, and na\"{i}vely scaling ridge regression results in overfitting.

The rest of this paper is set out as follows.  Section \ref{sec-background} considers relevant related work.  Section \ref{sec-method} sets out the details of the method.  Section \ref{sec-behaviour-large-p} proposes a connection between {\pv} and {\lr} for large $p$.  Section \ref{sec-experiments} presents the experimental results.  Section \ref{sec-discussion} reflects on the link between {\pv} and {\lr} in light of the results.  Section \ref{sec-limitations} considers the limitations of {\pv}.


\section{Background and Related Work} \label{sec-background}

\subsection{Logistic Regression}

Logistic regression directly models the probability of a given class as a linear combination of input features transformed through the softmax (logistic) function \citep{McCullaghNelder89,bishop_2006,hastie_etal_2009}. Let $\bX \in \mathbb{R}^{n \times p}$ denote an $n \times p$ design (feature) matrix with $n$ rows (examples) and $p$ columns (features), let $\by \in \left\{ 1, \ldots, L\right\}^n$ denote the vector of class labels associated with each observation, and let $\bm{\beta}^{(j)}$ and $\bm{\eta}^{(j)} = \bX\bm{\beta}^{(j)}$ denote the coefficients and linear predictors associated with class $j \in \{1,\ldots,L\}$, respectively. Then, under the logistic regression model
\begin{equation*}%
        \mathbb{P}(Y_i = j) = \frac{e^{\eta_i^{(j)}}}{\sum_k e^{\eta_i^{(k)}}} 
\end{equation*}%
is the predicted probability that observation $i$ is in class $j$.
The coefficients of logistic regression are typically chosen so as to minimise the sum of the negative log-likelihood (also called the ``cross-entropy loss'') plus a ridge or other regularisation penalty \citep{bishop_2006,hastie_etal_2009}. Let ${\bf H}(\bm{\beta}^{(1)},\ldots,\bm{\beta}^{(L)}) = (\bm{\eta}^{(1)},\ldots,\bm{\eta}^{(L)})$ denote the matrix of linear predictors given the set of coefficients; then the negative log-likelihood can be written as:%
\begin{equation}
    \label{eq:neg:log:like}
   l({\bf H}) = \sum_{i = 1}^{n} \left[ -\eta_{i}^{(y_{i})} + \log \sum_{j = 1}^{L} \exp \eta_{i}^{(j)} \right].  %
\end{equation}
and the $\ell_2$ (ridge) penalised maximum likelihood estimate is found by solving
\begin{equation}
    \hat{\bm{\beta}}_{\lambda} = \argmin_{\bm{\beta}^{(1)},\ldots,\bm{\beta}^{(L)}} \left\{ l({\bf H}(\bm{\beta}^{(1)},\ldots,\bm{\beta}^{(L)})) + \lambda \sum_{j=1}^{L} || \bm{\beta}^{(j)} ||^{2} \right\}, \label{eq-reg-ml}%
\end{equation}
where $||\cdot||^2$ denotes the squared $\ell_2$ norm. There is no closed-form solution to this minimisation problem.  Instead, the coefficients are usually found numerically by using some variant of gradient descent.  The choice of $q$ in an $\ell_{q}$ penalty reflects a prior in relation to the sparsity of the features, and the appropriateness of this prior will be problem dependent \citep[][pp 69--73]{hastie_etal_2009}.  Here, as we are making a direct comparison with a model based on linear regression with a ridge penalty, we concentrate on logistic regression with a ridge penalty, i.e., $q=2$.  For logistic regression, evaluating different candidates for the ridge penalty requires refitting the model, possibly multiple times for each candidate value of the ridge parameter (e.g.,~under~{$k$\nobreakdash-fold}~cross-validation).

\subsection{Ridge Regression}

In contrast to {\lr}, ridge regression models a numeric response as a linear combination of the input features \citep{hastie_etal_2009}: $\bm{\eta} = \bX\bm{\beta}_{\lambda}$.  The coefficients of a ridge regression model are chosen so as to minimise squared error between the predicted response and the target, plus a ridge penalty:%
$$\hat{\bm{\beta}}_{\lambda} = \argmin_{\bm{\beta}} \left\{ \lVert \by - \bX\bm{\beta} \rVert^{2} + \lambda \lVert \bm{\beta} \rVert^{2} \right\}.$$%
For a given value of $\lambda$, the closed form solution to this optimization problem is given by:%
\begin{equation}
    \label{eq-ridge-solution}
    \hat{\bm{\beta}}_{\lambda} = (\bX^{\T} \bX + \lambda {\bf I})^{-1} \bX^{\T} \by.
\end{equation}
Fitting the ridge regression model is particularly efficient, as: (i)~the complexity of the fit is proportional to $\min(n, p)$, that is, the minimum of the number of examples and number of features \citep[see][pp 659--660]{hastie_etal_2009}; and (ii)~the value of the ridge parameter, $\lambda$, can be determined efficiently by using the shortcut method for estimating the LOOCV error, which allows for the evaluation of multiple candidate values of $\lambda$ while only fitting the model once: see Section \ref{sec-method}.

Ridge regression can be used for classification by encoding the class as a regression target.  Accordingly, instead of a vector of regression responses, $\by$, for $k$ classes we create $k$ response vectors $\tilde{\by}^{(j)}$ in which the classes are encoded as binary values, e.g., $\{-1, +1\}$, in a one-versus-rest fashion (\citealp{bishop_2006}, pp~184--186; \citealp{hastie_etal_2009}, pp~103--106).  With multiple regression responses representing multiple classes, we regress $\bX$ on to each $\tilde{\by}^{(j)}$ separately.  Class predictions correspond to the class with the largest response, $\bm{c} = \argmax_{j} \{ \bX \hat{\bm{\beta}}^{(j)} \}$.

\subsection{Calibration, Stacking, and Prevalidation}

The predictions of (essentially) any classifier can be transformed into probabilities (e.g., using the softmax function), and `calibrated' to match a given probability distribution \citep{platt_1999,guo_etal_2017}.  This draws on at least three different strands of work which are closely related, but not well connected in the literature, namely: probability calibration, stacking, and prevalidation.

Calibration involves transforming the output of a model to match a given probability distribution.  In particular, temperature scaling involves dividing the output of a model by a scalar `temperature' parameter, and passing the scaled output through the softmax function (\citealp{guo_etal_2017}; see also, e.g., \citealp{niculescu-mizil_and_caruana_2005, kull_etal_2017, kull_etal_2019}).  Temperature scaling represents a simplification of Platt scaling \citep{platt_1999}.

Temperature scaling amounts to fitting a logistic regression model to the predictions of another model.  This is very similar to the idea of stacking \citep{wolpert_1992}, where the predictions of one or more models are used as the input for another model, fit using a separate validation set or cross-validation.  Where the second model is logistic regression, this is essentially equivalent to temperature scaling.

This is also similar to `prevalidation' \citep{tibshirani_and_efron_2002, hoefling_and_tibshirani_2008}, from the statistics literature, where a model is used to make predictions using cross-validation, and the `prevalidated' predictions from all cross-validation folds are used together to fit another model (e.g., logistic regression or another probabilistic model).

In each of these contexts, fitting an additional parameter or model to the predictions of a given model requires unbiased estimates for the relevant objective function, to avoid overfitting.  \citet{wolpert_1992}, \citet{platt_1999}, and \citet{tibshirani_and_efron_2002} all note the possibility of using leave-one-out cross-validation in this context.  However, to the best of our knowledge, the use of the shortcut estimate of LOOCV error for ridge regression in this context has not been explored in practice.

Ridge regression has at least two advantages in this context.  The closed-form solution allows for fitting the model efficiently which, in turn, allows for computing the prevalidated predictions efficiently, even when refitting the model multiple times under $k$-fold cross-validation.  More importantly, given the shortcut estimate of the LOOCV error, a simple rearrangement of terms allows for computing the corresponding LOOCV prevalidated predictions with negligible additional computational expense: see Section \ref{sec-method}, below.  This means, in effect, that the prevalidated predictions are already computed in the standard process of fitting the model.  Accordingly, the process of finding a scaling parameter for the coefficients of the model so as to minimise log-loss can be integrated into the existing procedure for fitting the model and evaluating different candidate values for the ridge parameter.

Existing work does not appear to take full advantage of this possibility, or to recognise how broadly applicable and effective the resulting prevalidated ridge regression model is, particularly for high-dimensional data.  \citet{kowalczyk_etal_2004} apply different estimates of LOOCV error for support vector machines and Platt scaling to both support vector machines and ridge regression.  However, \citet{kowalczyk_etal_2004} do not appear to use the shortcut estimate of LOOCV error for ridge regression, and use a fixed value for regularisation.  \citet{vancalster_etal_2007} use an estimate of LOOCV error to tune the hyperparameters of a least squares support vector machine, and then use Platt scaling, but using 5-fold cross-validation.  \citet{adankon_and_cheriet_2009} use an estimate of LOOCV error to fit the hyperparameters of a least squares support vector machine, transforming the outputs of the model via Platt scaling.  However, \citet{adankon_and_cheriet_2009} leave the scaling parameters fixed, and do not calibrate the resulting model.


\section{Method} \label{sec-method}

We find a scaling parameter, $\kappa$, for the coefficients of a ridge regression model which minimises log-loss for the set of {\em prevalidated predictions}, $\tilde{\bH}_{\lambda}$, corresponding to the estimated LOOCV predictions:
\begin{equation}%
    \label{eq-fit-c}%
    \{ \hat{\kappa}, \hat{\lambda} \} = \argmin_{\kappa,\lambda} \left\{ l(\kappa \cdot \tilde{\bH}_{\lambda} ) \right\}
\end{equation}%
where $l(\cdot)$ is given by (\ref{eq:neg:log:like}). The key aspect of this approach is the use of the prevalidated predictions, $\tilde{\bH}_{\lambda}$, from the ridge regression with ridge parameter $\lambda$. Here, $\tilde{\bH}_{\lambda}$ is an $n \times k$ matrix, with entries
\[
    [\tilde{\bH}_{\lambda}]_{i,j} = {\bf x}^T_i \hat{\bm{\beta}}^{(j)}_{\lambda,-i},
\]
where $\hat{\bm{\beta}}^{(j)}_{\lambda,-i}$ denotes the ridge regression solution for class $j$ and regularization parameter $\lambda$, when we drop observation $i$ from our data (i.e., ``leaving it out''). The formulation (\ref{eq-fit-c}) allows us to exploit the LOOCV shortcut tricks to efficiently compute the matrix of prevalidated predictors $\tilde{\bH}$. The scaling parameter $\kappa$ may then be found with only a nominal additional computational cost over the procedure for fitting a standard ridge regression classifier (i.e., without calibration). An additional advantage is that by using (\ref{eq-fit-c}) we are now choosing $\lambda$ to directly minimise a form of cross-validated log-loss, rather than squared-error as per a usual ridge regression classifier formulation; in fact, in Section~\ref{sec-behaviour-large-p} we show that (\ref{eq-fit-c}) is (essentially) equivalent to a full LOOCV using the exact penalised likelihood estimation approach, (\ref{eq-reg-ml}), under suitable conditions. Full pseudocode is provided in Appendix \ref{sec-appendix-pseudocode}.




\subsection{Prevalidated Predictions}

To solve (\ref{eq-fit-c}) we require the prevalidated predictors for each class. It is well known that the sum of the squared LOOCV errors for ridge regression can be computed efficiently without explicitly refitting by using the ``shortcut formula''. This allows fast implementation of the usual ridge regression classifier, which optimises the squared cross-validation errors. Our prevalidation approach, however, requires not the sum of the squared errors, but the leave-one-out predictions themselves. While these quantities are likely implicit in the standard derivations of the shortcut formula, their explicit representation does not seem to be widely known, so for completeness we provide the following result:

\begin{theorem}
    \label{thm:cv:prediction:err}
    Let $\hat{\bbeta}_\lambda$ be a ridge regression estimate on data $\bX$ and $\by$, and let ${\bf e}(\lambda) = \by - \bX \hat{\beta}_\lambda$ be the residuals of the ridge regression. Then the LOOCV prediction for observation $i$ is
    \begin{equation}
        \label{eq:cv:prediction:err}        
        {\bf x}_i^T \hat{\bbeta}_{\lambda,-i} = y_i - \frac{e_i(\lambda)}{1 - d_i(\lambda)}
    \end{equation}
    where $d_i(\lambda) = \bx_i^\T (\bX^\T \bX + \lambda \bI)^{-1} \bx_i$ is the $i$-th diagonal of the hat matrix $\bX (\bX^\T \bX + \lambda  \bI)^{-1} \bX^\T$ and $\hat{\bbeta}_{\lambda,-i}$ are the ridge regression estimates when observation $i$ is omitted from $\bX$ and $\by$.
\end{theorem}

\noindent The proof is deferred to Appendix~\ref{sec-appendix-proof-theorem}. From Theorem~\ref{thm:cv:prediction:err}, the entries of the prevalidated linear predictor matrix $\tilde{\bH}_{\lambda}$ may be found using
\[
    [ \tilde{\bH}_{\lambda} ]_{i,j} = \tilde{y}_i^{(j)} - (1-d_i(\lambda))^{-1} {e_i^{(j)}(\lambda)} \; \; {\rm where} \; \; {\bf e}^{(j)}(\lambda) = \tilde{\bf y}^{(j)} - \bX \hat{\bbeta}^{(j)}_\lambda.
\]
 $[ \tilde{\bH}_{\lambda} ]_{i,j}$ are the errors of the complete data ridge regression fits for each example $i$ and class $j$, and $\hat{\bbeta}^{(j)}_\lambda$ are the complete data ridge regression estimates for class $j$.

\subsection{SVD Pre-Processing}

Given a complete data ridge regression fit, Theorem~\ref{thm:cv:prediction:err} allows us to cheaply compute all $n$ LOOCV predictions. To efficiently evaluate multiple candidate values of the ridge penalty, $\lambda$, we need to be able to compute the complete data residuals and the diagonals of the hat matrix efficiently. We now describe how this may be done by exploiting the fact that the ridge solutions are preserved under orthogonal transformations. Following \citet{tew_etal_2023}, we let $r = {\rm min}(n,p)$ and $m = {\rm max}(n,p)$, and let $\bX = \bU \bSigma \bV^\T$ denote the compact singular value decomposition (SVD) of $\bX$, where $\bU \in \mathbb{R}^{n \times r}$ and $\bV \in \mathbb{R}^{n \times r}$ are semi-orthogonal matrices and $\bSigma \in \mathbb{R}^{r \times r}$ is a diagonal matrix of singular values, i.e., $\bSigma = {\rm diag}(\bs)$ where $\bs = (s_1,\ldots,s_r)$. Given an SVD, we may rotate the feature matrix $\bZ = \bX \bV$ so that $\bZ$ is orthogonal, i.e., $\bZ^\T \bZ = \Sigma^2$. This orthogonality allows us to compute the ridge regression coefficient estimates for class $j$, say $\hat{\balpha}^{(j)}_\lambda$, in $O(r)$ time via
\[
    {[\hat{\balpha}_\lambda]}_k^{(j)} = g_i^{(j)} (s_i^2 + \lambda)^{-1}, \; k = 1,\ldots,r,
\]
where ${\bf g}^{(j)} = {\bf Z}^\T \tilde{\by}^{(j)}$. Given $\hat{\balpha}^{(j)}_\lambda$, we may compute the residuals in $O(nr)$ time via
\[
    {\bf e}^{(j)} = \tilde{\by}^{(j)} - \bZ \hat{\balpha}_{\lambda}^{(j)}.
\]
The hat matrix for a ridge regression is invariant under orthonormal rotations of the feature space, and the orthogonality of $\bZ$ allows us to efficiently compute the diagonals of the hat matrix using (e.g., Equation (16) in \citet{tew_etal_2023})
\[
    d_i(\lambda) = \sum_{j=1}^r u^2_{i,j} s_j^2 / (s_j^2 + \lambda),
\]
so that all $r$ diagonal entries can be computed in $O(r^2)$ time. An important aspect of this procedure is that the SVD needs to be found only once, irrespective of the number of classes $L$ or number of different candidate values of $\lambda$ that we are evaluating. The SVD can be computed via eigendecomposition of $\bX^{\T} \bX$~(for~\smash{$n \geq p$}), or $\bX \bX^{\T}$~(for~\smash{$n < p$}) for a time complexity of $O(mr^2)$, i.e., the time complexity of a single least-squares fit. We also need to compute each ${\bf g}^{(j)}$ only once, irrespective of the number of candidate values of $\lambda$, and for each value of $\lambda$ we need to compute the hat-matrix diagonals $d_i(\lambda)$ only once irrespective of the number of classes. This makes the whole process extremely efficient.

\subsection{Scaling Parameter and Coefficients}

For each value of $\lambda$ we find the scaling parameter, $\kappa$, which minimises log-loss for the prevalidated predictions, $\tilde{\bH}_{\lambda}$, via gradient descent (using BFGS). We use the values of $\lambda$ and $\kappa$ which, together, minimise log-loss.  There are two differences between this approach, and the usual ridge regression classifier: (i) the value of $\lambda$ which minimises log-loss is not necessarily the same as the value of $\lambda$ which minimises squared error, and (ii) there is a shared value of $\lambda$ across all $L$ regressions.

Let $\hat{\kappa}$ and $\hat{\lambda}$ denote the values of $\kappa$ and $\lambda $ that minimise the prevalidated log-loss, i.e., the solutions of (\ref{eq-fit-c}). The final estimates of the coefficients for class $j$ in the original feature space are then given by
\[
    \hat{\bbeta}^{(j)} = \hat{\kappa} \bV \hat{\balpha}_{\hat{\lambda}}^{(j)}
\]
For feature vector $\bx$, the class probabilities, $\hat{P}_j(\bx)$, and class prediction, $\hat{c}(\bx)$, are given by
\[
    \hat{P}_j(\bx) = \frac{e^{\bx^\T \hat{\bbeta}^{(j)}}}{\sum_{k=1}^L e^{\bx^\T \hat{\bbeta}^{(k)}}} \; \; {\rm and} \; \; \hat{c}(\bx) = \argmax_{j} \hat{P}_{j}
\]
respectively. We note that if desired, it is also straightforward to compute the weights for each of the $n$ LOOCV models, and subsequently, for any given example, the predictions corresponding to each of the separate LOOCV models: see Appendix \ref{sec-appendix-loocv}.

Note also that while we optimise the scaling parameter via gradient descent, in contrast to (maximum likelihood) logistic regression, we are optimising only a single scalar parameter, $\kappa$, rather than $p \times k$ coefficients.  In principle, it is possible to jointly optimise $\lambda$ and $\kappa$ via gradient descent.  However, the LOOCV loss is not necessarily convex with respect to $\lambda$, potentially having multiple poor local minima \citep{stephenson_etal_2021,tew_etal_2023}, so instead we resort to a grid search for $\lambda$.

We implement {\pv} in Python, using \texttt{numpy.linalg.eigh} for the eigendecomposition, and \texttt{scipy.optimise.minimize} to fit the scaling parameter.  Our code is available at: \url{https://github.com/angus924/preval}.

\section{Behaviour of Prevalidated Ridge Classifier for Large $p$} \label{sec-behaviour-large-p}

We will now argue that minimising the prevalidated log-loss should approximate estimation via exact LOOCV penalised maximum likelihood ridge regression, at least for large $p$. For clarity of exposition we restrict ourselves to the case of binomial logistic regression, i.e., $L=2$. The arguments here generalise to multiple classes with minor modifications. We also assume the class labels are $y_i = 0$ and $y_i = 1$ to keep the mathematics slightly simpler, as the particular choices have no impact in practice. The iteratively reweighted least squares (IRLS) algorithm is a standard algorithm for fitting GLMs via penalised maximum likelihood~\citep{McCullaghNelder89}. Assume that the columns of $\bX$ are centred, and that $\bX$ is augmented with an additional column of ones to model the intercept. For ridge penalised logistic regressions each iteration of this algorithm updates the coefficients $\bbeta_t$ using
\begin{equation}
    \label{eq:IRLS:update}
    \bbeta^{t+1} = \left( \bX^\T \bW^t \bX + \lambda \bA \right)^{-1} \bX^\T \bW^{t} \bz^t
\end{equation}
where $\bA$ is a $(p+1) \times (p+1)$ matrix with a $p \times p$ identity matrix in the top left and zeros everywhere else, 
\[
    z^t_{i} = \bx_i^\T \bbeta^t +  \frac{y_i - \phi(\bx_i^\T \bbeta^t)}{v^t_i} \; \; {\rm and} \; \; \bW^t = {\rm diag}({\bf v}^t)
\]
are the rescaled targets and weights at iteration $t$, $\phi(x) = e^x/(1+e^x)$ is the logistic transformation and the vector of weights is given by 
\[
    v^t_{i} = \phi(\bx_i^\T \bbeta^t) \left( 1 - \phi(\bx_i^\T \bbeta^t) \right).
\]
The entries of $\bv^t$ are the predicted variances for each observation at step $t$ under the binomial model. We have the following proposition.

\begin{theorem}
    \label{thm:prop:1}
    Let $e_i = y_i - \phi(\bx_i^\T \bbeta^t)$ and let $\hat{\beta}^{\rm RR}_\lambda$ denote the solution to the usual ridge regression optimisation problem (\ref{eq-ridge-solution}) with regularisation parameter $\lambda$. Then, if $|e_1| = \cdots = |e_n| = \varepsilon$ the solution to (\ref{eq:IRLS:update}) at iteration $t+1$ is exactly equal to $c(\varepsilon) \, \hat{\beta}^{\rm RR}_{\lambda_*}$, where $\lambda_* = \varepsilon^{-1} (1-\varepsilon)^{-1} \lambda$ and 
    \[
        c(\varepsilon) = 2 \, \log \left(\frac{1-\varepsilon}{\varepsilon} \right) + \frac{2}{1-\varepsilon}.
    \]
\end{theorem}

\noindent The proof of Theorem~\ref{thm:prop:1} is deferred to Appendix~\ref{sec-appendix-large-p}. Theorem \ref{thm:prop:1} tells us that if a solution $\bbeta_t$ to an IRLS iteration fits the data $\by$ with equal error everywhere, then the solution to the update $\bbeta_{t+1}$ will be exactly equal to a scaled version of a solution to the usual squared-error ridge regression problem (\ref{eq-ridge-solution}) with an appropriate regularisation parameter. As an important example, if we initialise $\bbeta_0 = {\bf 0}$ then in the first iteration of IRLS, $\phi(\bx_i^\T \bbeta_0) = 1/2$ for all $i$, so $\varepsilon_0 = 1/2$, and $\bbeta_1 = 4 \, \hat{\bbeta}(4 \lambda)$, i.e., the first step in an IRLS will always be exactly equal to a scaled version of a ridge solution with $\lambda_* = 4\lambda$. Therefore, by Theorem~\ref{thm:prop:1}, the ridge regression solution given by (\ref{eq-ridge-solution}) can be viewed as a form of one-step IRLS estimator. We note this holds for all $p$ and $\lambda>0$. 


We will now argue that if $p \gg n$ and $\lambda$ is not too large the solutions of multiple IRLS steps will also be (approximately) scalars of appropriate ridge regression solutions. To see this, we note that if $p \gg n$, the problem is underdetermined and we can fit $\by$ as closely as we desire by taking $\lambda$ to be as small as necessary. Therefore the solution after the first step, $\bbeta_1$, will closely fit $\by$ and from the properties of the squared-error measure, most errors $e_i$ will be of similar magnitude, say $\varepsilon_1$, i.e., $|e_i| \approx \varepsilon_1$ for all $i$. Then, we may apply Theorem \ref{thm:prop:1} again to show that the next step $\bbeta_2$ will also be a scaled version of a ridge regression solution, with a larger scalar (as $\varepsilon_1 < 1/2$, $1/2$ being the value of $\varepsilon$ in the first step of the algorithm); this solution will also fit $\by$ closely and most errors will be of similar size, say $\varepsilon_2 < \varepsilon_1$, and the argument may be repeated. After a certain number of iterations, say $T$, the algorithm will converge; the size of the error in fitting $\by$ at step $T$, $\varepsilon_T$, will be determined by the sample size $n$ and the regularisation parameter $\lambda$. The larger $n$ the smaller $\varepsilon_T$ will be for a given $\lambda$, as the ridge estimator will be allowed to fit the $\by$ more closely; therefore $\varepsilon_T \equiv \varepsilon_T(n,\lambda)$ and from Theorem \ref{thm:prop:1} we have
\begin{eqnarray*}
    \bbeta_T &\approx& c\left(\varepsilon_T(n,\lambda)\right) \hat{\bbeta}^{\rm RR}_{\lambda_*} \\
    &=& \kappa(n,\lambda) \hat{\bbeta}^{\rm RR}_{\lambda_*}
\end{eqnarray*}
where $\lambda_* = \varepsilon_T(n,\lambda)^{-1} (1-\varepsilon_T(n,\lambda))^{-1} \lambda$.
%
%
Let $L(y, \theta)$ denote the cross-entropy log-loss for binary observation $y$ with estimated probability that $y=1$ given by $\theta$. Then, there is a value of $\lambda_*$ and $\kappa_*$ such that the exact penalised ML LOOCV log-loss satisfies 
\begin{eqnarray*}
    \nonumber
    \sum_{i=1}^n L\left(y_i, \phi\left(\bx_i^\T \hat{\beta}^{\rm ML}_{\lambda,-i}\right)\right) &\approx& \sum_{i=1}^n L\left(y_i, \phi\left(\kappa(\lambda,n-1) \bx_i^\T \hat{\beta}^{\rm RR}_{\lambda_*,-i}\right)\right)\\
    &\approx& \sum_{i=1}^n L\left(y_i, \phi\left(\kappa_* \, \bx_i^\T \hat{\beta}^{\rm RR}_{\lambda_*,-i}\right)\right),
\end{eqnarray*}
where $\phi(x) = e^x/(1+e^x)$ is the logistic transformation. The right-hand-side of the above equation is exactly the prevalidated log-loss minimised in (\ref{eq-fit-c}); hence, for large $p$ minimizing the prevalidated log-loss approximates minimizing the exact penalised ML LOOCV log-loss.  The experiments show that {\pv} closely matches {\lr} in terms of both {\zo} and log-loss for large~$p$.


\section{Experiments} \label{sec-experiments}

We evaluate {\pv} and {\lr} (with ridge penalty, i.e., fitted by solving (\ref{eq-reg-ml})) on 273 datasets, drawn from four domains---tabular data, microarray data, image data, and time series data---demonstrating that {\pv} closely matches both the {\zo} and log-loss of {\lr}, while being significantly more computationally efficient.  Full details for each experiment as well as a sensitivity analysis are provided in the Appendix.

\subsection{Tabular Data}

\begin{figure}%
    \centering%
    \includegraphics[width=\linewidth]{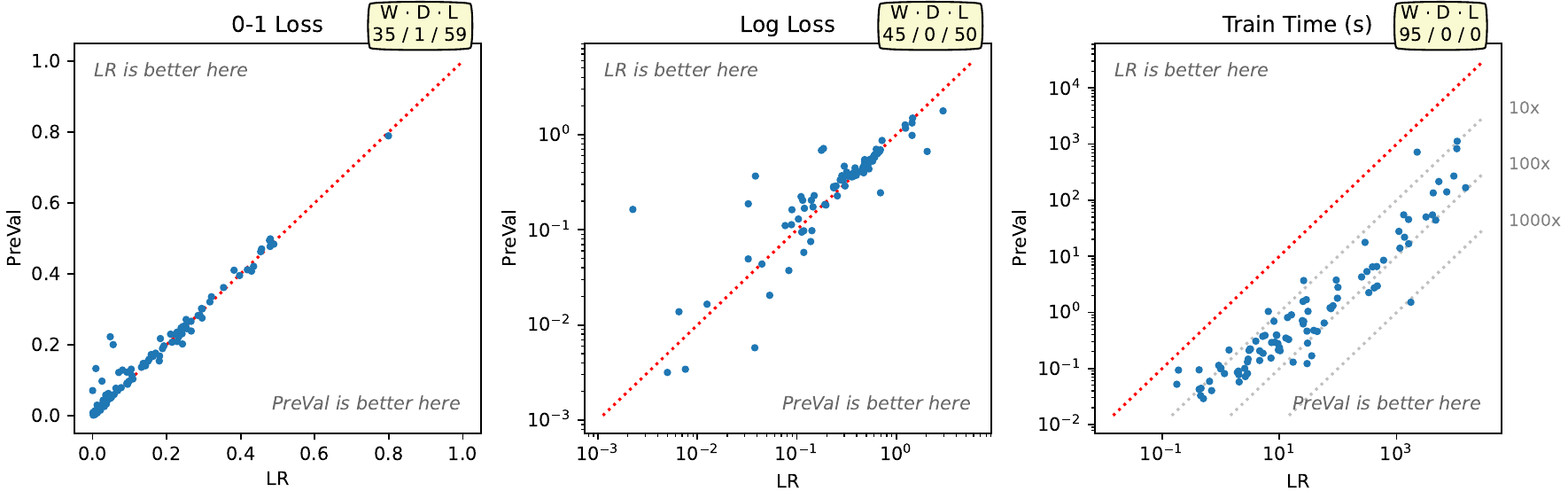}%
    \vspace{-1em}%
    \caption{Pairwise {\zo} (left), log-loss (centre), and training time (right) for {\pv} vs {\lr} on tabular datasets (with interactions).  {\pv} closely matches {\lr} for both \mbox{0--\!1} and log-loss on most datasets while requiring a fraction of the training time.}%
    \label{fig-tabular-interactions}%
\end{figure}%

We evaluate {\pv} and {\lr} in terms of {\zo}, log-loss, and training time on 95 datasets from the OpenML repository, consisting of 72 datasets from the OpenML CC18 benchmark \citep{bischl_etal_2021}, as well as 23 datasets from \citet{grinsztajn_etal_2022}.  Most of these datasets are `large $n$, small $p$', with between $500$ and $940{,}160$ examples, and between $4$ and $3{,}113$ features.

Figure \ref{fig-tabular-interactions} shows that, for the original features plus the interactions, {\pv} closely matches {\lr} in terms of both {\zo} (left) and log-loss (centre) on most datasets, while requiring significantly less training time (right).  {\pv} has lower {\zo} than {\lr} on 35 datasets, while {\lr} has lower {\zo} than {\pv} on 59 datasets.  {\pv} has lower log-loss on 45 datasets, while {\lr} has lower log-loss on 50 datasets.  The median advantage of {\pv} in terms of training time is approximately $30\times$.  Figure \ref{fig-tabular-original} (Appendix) shows that, for the original features (in most cases, $n \gg p$), {\lr} has a greater advantage over {\pv} in terms of both {\zo} and log-loss, although the differences are small on most datasets, and {\pv} has a smaller advantage over {\lr} in terms of training time.

\subsection{Microarray Data}

\begin{figure}%
    \centering%
    \includegraphics[width=\linewidth]{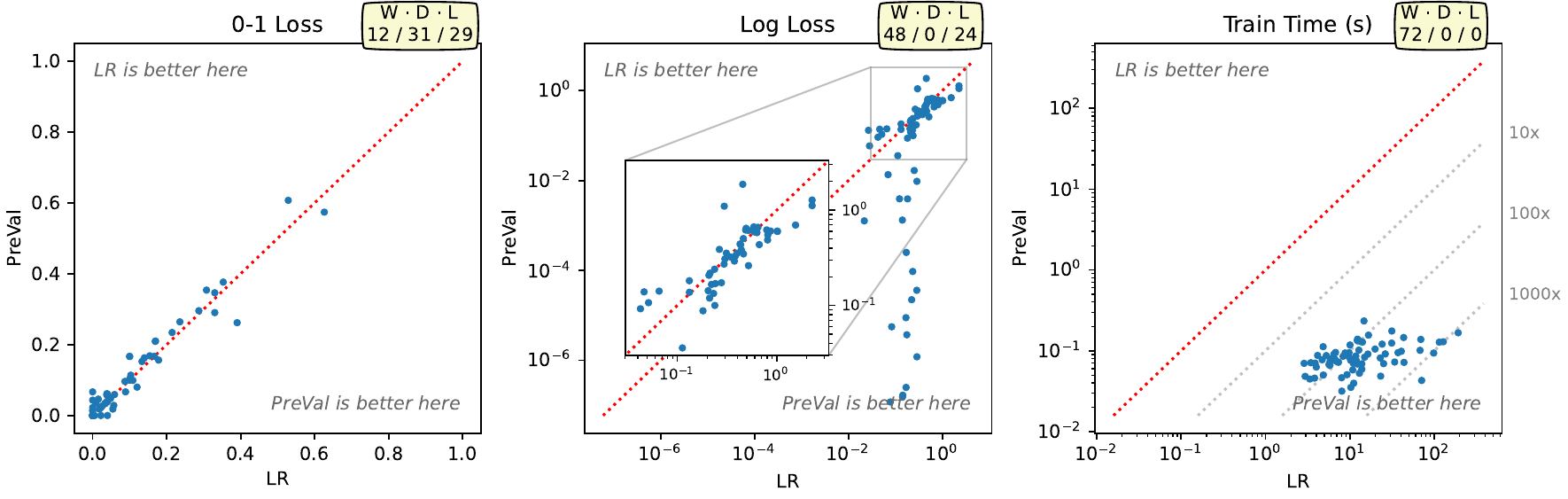}%
    \vspace{-1em}%
    \caption{Pairwise {\zo} (left), log-loss (centre), and training time (right) for {\pv} vs {\lr} on microarray datasets.  {\pv} closely matches {\lr} in terms of both \mbox{0--\!1} and log-loss while requiring only a small fraction of the training time.}%
    \label{fig-microarray}%
\end{figure}%

We evaluate {\pv} and {\lr} on 72 microarray classification datasets from the {CuMiDa} benchmark \citep{feltes_etal_2019}.  These are all `large $p$, small $n$' datasets, with between $12{,}621$ and $54{,}676$ features, and between $12$ and $357$ examples.

Figure \ref{fig-microarray} shows that {\pv} closely matches {\lr} in terms of both {\zo} (left) and log-loss (centre) on most datasets, while requiring up to $1000\times$ less training time (right).  {\pv} has lower {\zo} on 12 datasets, and {\lr} has lower {\zo} on 29 datasets (both models produce the same {\zo} on 31 datasets).  {\pv} has lower log-loss on 48 datasets (multiple orders of magnitude lower on several datasets), and {\lr} has lower log-loss on 24 datasets.  The median advantage of {\pv} over {\lr} in terms of training time is approximately~$140\times$.

\subsection{Image Data} \label{subsec-image-data}

\begin{figure}%
    \centering%
    \includegraphics[width=\linewidth]{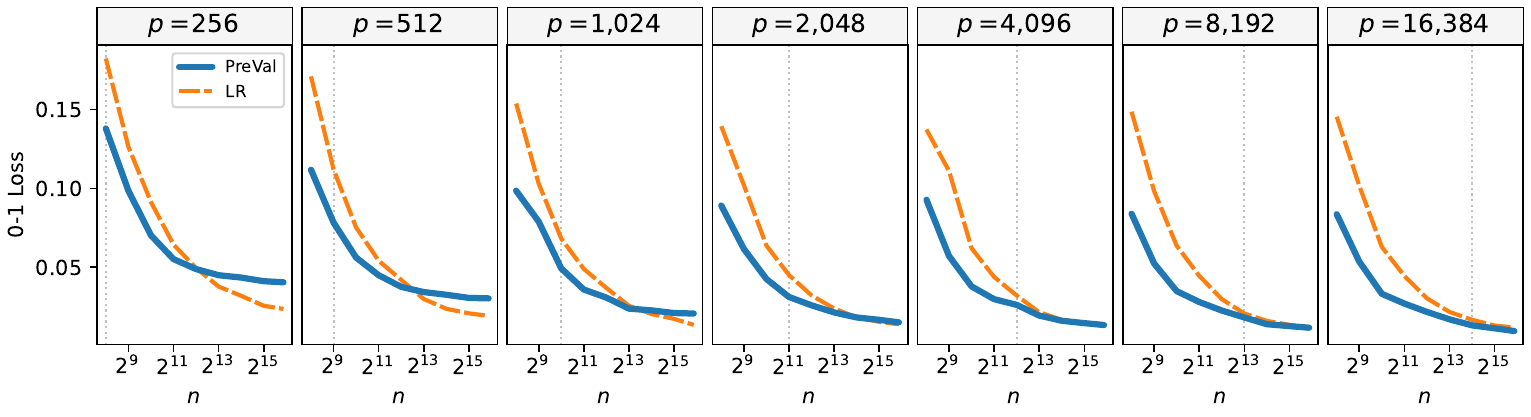}%
    \vspace{-1em}%
    \caption{Learning curves ({\zo}) for {\pv} (blue) and {\lr} (orange) for a random projection of the MNIST dataset.  {\pv} achieves consistently lower {\zo} for small $n$.  Asymptotic {\zo} for {\pv} approaches that of {\lr} as $p$ grows.}%
    \label{fig-mnist-projection-accuracy}%
\end{figure}%

We evaluate {\pv} and {\lr} on both the MNIST \citep{lecun_etal_1998} and Fashion-MNIST \citep{xiao_etal_2017} datasets.  For each dataset, we evaluate {\pv} and {\lr} on both the original features (pixels), as well as on a random projection with between $p = 256$ and $p = 16{,}384$ features.

Figure \ref{fig-mnist-projection-accuracy} shows learning curves ({\zo}) for {\pv} and {\lr}, for increasing numbers of features, $\smash{p~\in~\{2^{8},~2^{9},~\ldots,~2^{14}\}}$, for a random projection of the MNIST dataset.  Figure \ref{fig-mnist-projection-loss} (page \pageref{fig-mnist-projection-loss}) shows the corresponding learning curves for log-loss.  Figures \ref{fig-mnist-projection-loss} and \ref{fig-mnist-projection-accuracy} show that while the {\zo} and log-loss of both models improves as the number of features increases, both the {\zo} and log-loss of {\pv} improves relative to {\lr}.  For a small number of features, the asymptotic {\zo} and log-loss of {\lr} is superior to that of {\pv}.  However, as the number of features increases, the {\zo} and log-loss of {\pv} matches or betters that of {\lr}.  Figures \ref{fig-fashion-projection-loss} and \ref{fig-fashion-projection-accuracy} (Appendix) show similar results for a random projection of the Fashion-MNIST dataset.  Figures \ref{fig-mnist-all} and \ref{fig-fashion-mnist-all} (Appendix) show that for the original features, {\lr} achieves lower {\zo} and log-loss as compared to {\pv} regardless of training set size.  Figure \ref{fig-fashion-mnist-projection-time} (Appendix) shows training times for {\pv} and {\lr} for the random projections of the MNIST and Fashion-MNIST datasets.  Unsurprisingly, {\pv} is significantly more efficient than {\lr}.  {\pv} is approximately $16\times$ faster than {\lr} in terms of training time for ${n=60{,}000}$ and ${p=16{,}384}$.

\subsection{Time Series Data}

\begin{figure}%
    \centering%
    \includegraphics[width=\linewidth]{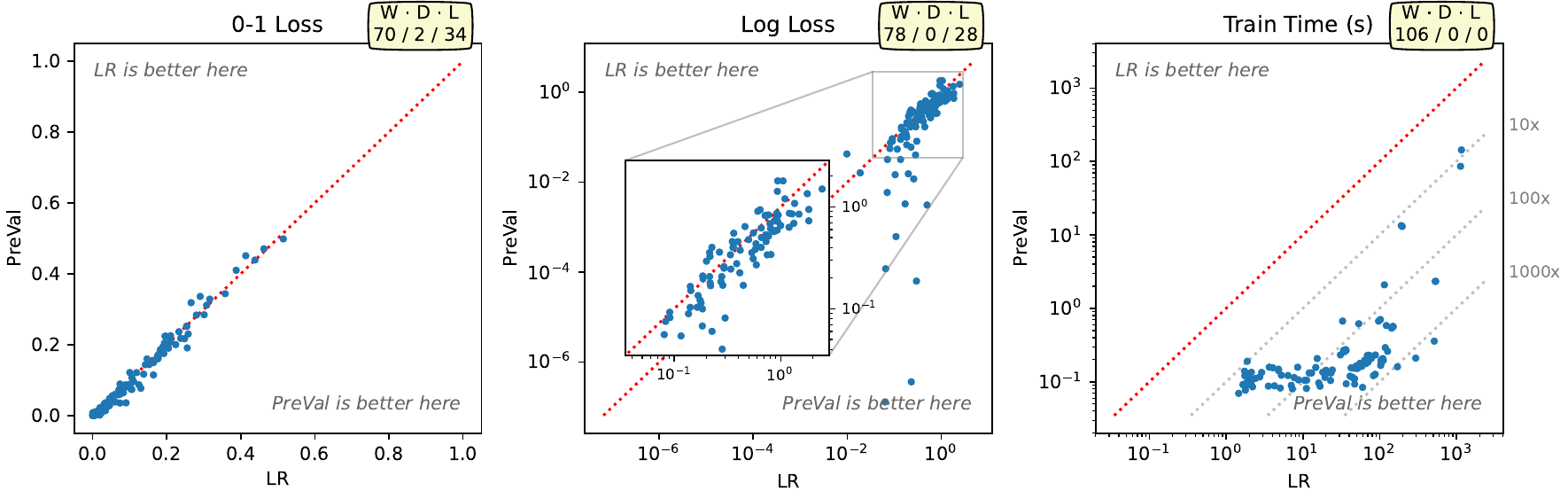}%
    \vspace{-1em}%
    \caption{Pairwise {\zo} (left), log-loss (centre), and training time (right) for {\pv} vs {\lr} on time series datasets.  {\pv} closely matches {\lr} in terms of both \mbox{0--\!1} and log-loss while requiring only a fraction of the training time.}%
    \label{fig-ucr}%
\end{figure}%

We evaluate {\pv} and {\lr} on 106 datasets from the UCR archive \citep{dau_etal_2019}.  For each dataset, we transform the time series using the popular {\minirocket} transform \citep{dempster_etal_2021}.  The transformed datasets are almost all `large $p$, small $n$', with approximately $10{,}000$ features (from the {\minirocket} transform), and between $17$ and $8{,}926$ training examples.

Figure \ref{fig-ucr} shows that, again, {\pv} achieves similar {\zo} (left) and log-loss (centre) to {\lr} on most datasets, while being between approximately $10\times$ and $1{,}000\times$ faster to train (right).  {\pv} has lower {\zo} on 70 datasets, and {\lr} has lower {\zo} on 34 datasets.  {\pv} has lower log-loss on 78 datasets (multiple orders of magnitude lower on a small number of datasets), while {\lr} achieves lower log-loss on 28 datasets.  The median advantage of {\pv} over {\lr} in terms of training time is approximately $140\times$.


\section{Discussion} \label{sec-discussion}

Consistent with the argument presented in Section \ref{sec-behaviour-large-p}, the experiments show that, in the high-dimensional regime, {\pv} closely matches {\lr} in terms of both {\zo} and log-loss.  Scaling the parameters of the ridge regression model so as to minimise log loss in effect transforms the ridge regression model into a logistic regression model.  However, while {\pv} and {\lr} are both in some sense logistic regression models, they are different.

In particular, {\pv} is restricted to the subspace of models which are scalar multiples of ${\hat{\bm{\beta}}_{\lambda} = (\bf{X}^{\T} \bf{X} + \lambda \bf{I})^{-1} \bf{X}^{\T} \bf{Y}}$, i.e., the linear ridge regression solution, whereas a logistic regression model fitted via regularised maximum likelihood is not.  Nevertheless, as the size of the feature space grows, the experimental results show that a model within the subspace of the ridge solution can fit the data as well as {\lr}, providing empirical support the proposition that minimising prevalidated log-loss for the ridge regression model increasingly closely approximates exact LOOCV penalised maximum likelihood logistic regression for a sufficiently-large feature space.


\section{Limitations} \label{sec-limitations}

While {\pv} has several advantages over maximum likelihood logistic regression, {\pv} also has a number of limitations.  {\pv} is most effective in the high-dimensional regime.  This is where {\pv} most closely matches {\lr} in terms of {\zo} and log-loss, and has the greatest advantage over {\lr} in terms of computational efficiency.  {\lr} is more effective for low-dimensional datasets.

{\pv} is most effective for datasets with dense, rather than sparse, features.  {\pv} is based on ridge regression.  In this sense, {\pv} assumes that ridge regularisation represents an appropriate prior in relation to the sparsity of the features.  (This is the same assumption made by {\lr} fitted using $\ell_{2}$ penalised maximum likelihood.)  Models using, e.g., $\ell_{1}$ regularisation are likely to be more effective for datasets with sparse features.

{\pv}, as currently implemented, is intended to be used where the dataset fits in memory, and where the relevant computations (e.g., SVD) can be conducted in memory.  While the experiments demonstrate that this is possible for datasets with at least $60{,}000$ training examples and $16{,}384$ features, for larger datasets (which do not fit in memory), optimising the coefficients via gradient descent would be more appropriate.  We leave this for future work.


\section{Conclusion}

We demonstrate that {\pv}, a ridge regression model with its coefficients scaled so as to minimise log-loss for the set of prevalidated predictions derived from LOOCV error, closely matches regularised maximum likelihood logistic regression in terms of both {\zo} and log-loss, while incurring significantly lower computational cost.  The resulting model yields probabilistic predictions which closely match those of {\lr}, but with nominal additional computational expense over a standard ridge regression model.

{\pv} is particularly effective in the high-dimensional regime.  This is relevant to a wide variety of contexts: high-dimensional tabular data (including tabular data with a large number of interactions), high-dimensional biomedical data (such as microarray or other genomic data), and various high-dimensional transformations or projections of unstructured data such as image data (e.g., the features produced by a convolutional neural network) and time series data (e.g., the features produced by one of the many high-dimensional transforms commonly used in this context).  While {\pv} and {\lr} are different models, and one model should not be expected to achieve lower {\zo} or log-loss than the other on all datasets, we demonstrate that {\pv} is very attractive as a drop-in replacement for {\lr} in many contexts, particularly for high-dimensional data, while requiring significantly lower computational cost.

\bibliography{references}

\newpage
\appendix


\section{Appendix}


\subsection{Experimental Details}

We use the scikit-learn implementation of {\lr} with its default hyperparameter values: ridge regularisation, internal 5-fold cross-validation, and fit using LBFGS.  (We leave the maximum number of iterations for LBFGS at its default value.  We find that increasing the maximum number of iterations makes little practical difference to classification error or log-loss.  Any improvements due to a greater number of iterations, however modest, would come at proportionally greater computational expense.)  Unless otherwise specified, we normalise the input features by subtracting the mean and dividing by the standard deviation as computed on the training set.  All experiments are conducted on a cluster using a mixture of Intel Xeon and Intel Xeon Gold CPUs, using one CPU core per dataset per run.  Total runtime (all models including logistic regression) for the experiments was approximately 88 hours (approx. 4 days): approx. 30 minutes for the microarray datasets, approx. 2 hours 30 minutes for the time series datasets, approx. 28 hours for the tabular datasets, approx. 57 hours for the image datasets.  Maximum memory requirement was 64 GB (although most experiments can be conducted with 16 GB).

\subsubsection{Tabular Data}

We evaluate {\pv} and {\lr} using 5-fold stratified cross-validation (for each fold, $80\%$ of the data is used for training, and $20\%$ of the data is used for evaluation).  We use one-hot encoding for categorical features.  We evaluate {\pv} and {\lr} on both the original features, as well as the original features plus the order-2 interactions.  We limit the maximum in-memory size of each dataset to $4$~GB using float32 precision.  For datasets where the addition of the interactions would require more than $4$~GB, we use a random subset of interactions.  Additionally, we note that the CC18 benchmark includes the MNIST and Fashion-MNIST datasets, which are also evaluated separately in Section \ref{subsec-image-data}.

\subsubsection{Microarray Data}

We evaluate {\pv} and {\lr} using 5-fold stratified cross-validation.  Due to the distribution of the features and the small number of examples, we normalise the data by subtracting the median.

\subsubsection{Image Data}

We use the default split provided for each dataset ($60{,}000$ training examples, $10{,}000$ test examples).  For the random projection, each feature is computed by convolving a given input image with a random convolutional kernel (with a height and width of 9 pixels, and with weights drawn from a unit normal distribution, and no bias), and performing global average pooling (after ReLU).  We note that in practice, especially for very large training set sizes, it may be more efficient to train logistic regression (and the prevalidated model) via gradient descent using a single holdout validation set.

\subsubsection{Time Series Data}

We use the subset of 106 of the 112 datasets used in \citet{middlehurst_etal_2021} where there are at least 5 training examples per class.  Following standard practice for this benchmark, we evaluate {\pv} and {\lr} on 30 resamples of each dataset, using the same resamples as in \citet{middlehurst_etal_2024}.

\clearpage


\subsection{Sensitivity Analysis} \label{sec-sensitivity}

We highlight the effect of two key aspects of the method, namely: (a)~the scaling parameter; and (b)~the use of the prevalidated predictions.

\subsubsection{Scaling Parameter}

Figure \ref{fig-mnist-projection-loss} (page \pageref{fig-mnist-projection-loss}) shows learning curves (log-loss) for {\pv} and {\lr}, as well as for a standard ridge regression model without scaling (green), for a random projection of the MNIST dataset.  (For the standard, unscaled ridge regression model, we use the ridge regression implementation from scikit-learn.)  The predictions of the ridge regression model have been `squashed' through the softmax function.  Figure \ref{fig-fashion-projection-loss} (Appendix) shows equivalent results for a random projection of the Fashion-MNIST dataset.  Figures~\ref{fig-mnist-projection-loss} and \ref{fig-fashion-projection-loss} show that, without scaling, the outputs of a ridge regression model produce poor (high) log-loss compared to {\lr} or {\pv}, regardless of the number of features or the number of training examples.

\subsubsection{Prevalidated Predictions}

\begin{figure}[!h]%
    \centering%
    \includegraphics[width=\linewidth]{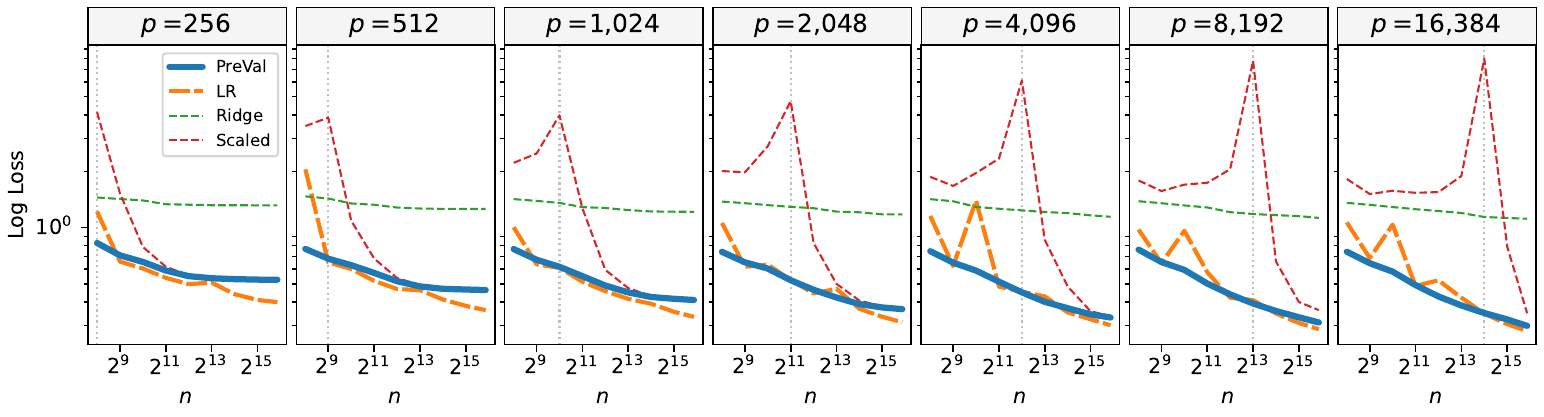}%
    \vspace{-1em}%
    \caption{Learning curves (log-loss) for {\pv} (blue), {\lr} (orange), ridge regression (green), and na\"{i}vely scaled ridge regression~(red), for increasing numbers of features, $p \in \{2^{8}, 2^{9}, \ldots, 2^{14}\}$, for a random projection of the Fashion-MNIST dataset.  {\pv} closely matches {\lr} in terms of log-loss in most scenarios while requiring a fraction of the computational expense.  Corresponding results for the MNIST dataset are shown in Figure \ref{fig-mnist-projection-loss}.}%
    \label{fig-fashion-projection-loss}%
\end{figure}

Figure \ref{fig-mnist-projection-loss} also shows learning curves for a ridge regression model where the scaling parameter has been tuned `na\"{i}vely' using unvalidated predictions (red), i.e., for a model fit using all of the training data: see Section \ref{sec-method}.  Again, Figure \ref{fig-fashion-projection-loss} (Appendix) shows equivalent results for a random projection of the Fashion-MNIST dataset.

Figures \ref{fig-mnist-projection-loss} and \ref{fig-fashion-projection-loss} show that using unvalidated predictions produces poor (high) log-loss.  Using unvalidated predictions, the effect of the scaling parameter is to catastrophically overfit log-loss on the training set, in effect `undoing' the regularisation otherwise provided by the ridge penalty.  The appearance of the learning curves is reminiscent of double descent \citep[see generally][]{hastie_etal_2022}.  Maximum log-loss occurs at $p \approx n$.  While log-loss decreases as $n$ increases relative to $p$ (for the overdetermined regime where $n > p$), log-loss for the unvalidated predictions does not match log-loss for the prevalidated model until approximately $n \geq 8\times p$.  In other words, the number of training examples must far exceed the number of features before being sufficient to overcome the effects of overfitting.

\clearpage


\subsection{Pseudocode} \label{sec-appendix-pseudocode}

\begin{algorithm}
    \caption{{\pv}}
    \label{alg-preval}
    \begin{algorithmic}
        \REQUIRE $\bX$ : $n \times p$ design matrix (centred)
        \REQUIRE $\tilde{\by}$ : $n \times L$ one-vs-rest encoded target classes
        \IF{$n \geq p$}
        \STATE $\bm{\Sigma}^{2}, \bV \leftarrow \text{eig}(\bX^{\T} \bX)$
        \STATE $\bU \leftarrow \bX \bV \bm{\Sigma}^{-1}$
        \ELSE
        \STATE $\bm{\Sigma}^{2}, \bU \leftarrow \text{eig}(\bX \bX^{\T})$
        \STATE $\bV \leftarrow \bX^{\T} \bU \bm{\Sigma}^{-1}$
        \ENDIF
        \STATE $m^{*} \leftarrow \infty$ \hfill \COMMENT{initialise log-loss w.r.t. $c$ and $\lambda$}
        \STATE $\bZ \leftarrow \bU\bm{\Sigma}$
        \STATE ${\bf G} \leftarrow \bZ^{\T}\tilde{\by}$
        \FOR{$\lambda \in \{10^{-3}, \ldots, 10^{3}\}$}
        \STATE $\hat{\bm{\alpha}}_{\lambda} \leftarrow {\bf G} \big/ \left( \bm{\Sigma}^{2} + \lambda \right)$
        \STATE $\bH_{\lambda} \leftarrow \bZ \hat{\bm{\alpha}}_{\lambda}$ \hfill \COMMENT{predictions for `full' fit with all $\bX$}
        \STATE $\bm{e}_{\lambda} \leftarrow \tilde{\by} - \bH_{\lambda}$ \hfill \COMMENT{residuals for `full' fit}
        \STATE $d_{i} (\lambda) \leftarrow \sum_{j} \left[ \bZ^{2} \big/ (\bm{\Sigma}^{2} + \lambda) \right]_{.,j}$ \hfill \COMMENT{sum across rows}
        \STATE $\tilde{\bm{e}}_{\lambda} \leftarrow \frac{\bm{e}}{1 - d(\lambda)}$ \hfill \COMMENT{shortcut LOOCV residuals}
        \STATE $\tilde{\bH}_{\lambda} \leftarrow \bH_{\lambda} - (\tilde{\bm{e}}_{\lambda} - \bm{e}_{\lambda})$ \hfill \COMMENT{\textbf{prevalidated predictions}}
        \STATE $m, \kappa \leftarrow \text{minimise}(l(\kappa \cdot \tilde{\bH}))$ \hfill \COMMENT{via BFGS}
        \IF{$m < m^{*}$}
        \STATE $m^{*}, \hat{\bm{\alpha}}_{\hat{\lambda}}, \hat{\kappa} \leftarrow m, \hat{\bm{\alpha}}_{\lambda}, \kappa$
        \ENDIF
        \ENDFOR
        \STATE $\hat{\bm{\beta}} \leftarrow \hat{\kappa} \, \bV \! \hat{\bm{\alpha}}_{\hat{\lambda}}$
    \end{algorithmic}
\end{algorithm}

\subsection{Proof of Theorem~\ref{thm:cv:prediction:err}} \label{sec-appendix-proof-theorem}

Without any loss of generality, we will assume the columns of $\bX$ are centred, and for convenience we suppress the explicit dependence of $\hbbeta_\lambda$ on $\lambda$. To remove the explicit ridge penalization term from the objective function we adopt the standard representation of ridge regression as least-squares on a new design matrix and target vector augmented by $\sqrt{\lambda} {\bf I}_p$ and ${\bf 0}_p$, respectively. Let $\bX_{t}$ and $\by_t$ denote the submatrix and subvector made from the first $t$ observations of $\bX$ and $\by$, respectively. We the have the following lemma.

\begin{lemma}
    Let $\hbbeta_t$ denote the least-squares estimates for the first $t$ observations of $\bX$ and $\by$. Then, $\hbbeta_{t+1}$ is given by the following recurrence relation
    \begin{equation}
        \label{eq:ls:update}
        \hbbeta_{t+1} = \hbbeta_t + z_{t+1} \left( \bX_{t+1}^\T \bX_{t+1} \right)^{-1} \bx_{t+1}
    \end{equation}
    where $z_{t+1} = y_{t+1} - \bx_{t+1}^\T \hbbeta_{t}$ is the error in predicting $y_{t+1}$ using the least-squares estimates based on the first $t$ data points only.
\end{lemma}

\begin{proof}
    First note that the residual sum-of-squares on the complete data for a given $\bbeta$ is
    \begin{equation}
        \label{eq:rss:0}
        ||\by_{t+1} - \bX_{t+1} \bbeta||^2 = ||\by_{t} - \bX_{t} \bbeta||^2 + (y_{t+1} - \bx_{t+1}^\T \bbeta)^2.
    \end{equation}
    As the sum-of-squared residuals is a quadratic function it can be written as
    \begin{equation}
        \label{eq:rss:1}
        ||\by_{t} - \bX_{t} \bbeta||^2 = || \by_{t} - \bX \hbbeta_{t}||^2 + (\hbbeta_t - \bbeta)^\T \left( \bX_t^\T \bX \right) (\hbbeta_t - \bbeta)         
    \end{equation}
    where $\hbbeta_t$ is the least-squares estimate for $\bX_t$ and $\by_t$, i.e., the value of $\bbeta$ that minimises $||\by_{t} - \bX_{t} \bbeta||^2$. Defining $\bbeta = \hbbeta_t + \tbbeta$, and noting that the first term on the right-hand-side of (\ref{eq:rss:1}) is independent of $\bbeta$, we have
    \begin{equation}
        \label{eq:rss:2}
        ||\by_{t} - \bX_{t} \bbeta||^2 = c + \tbbeta^\T \left( \bX_t^\T \bX \right) \tbbeta
    \end{equation}
    where $c$ are constants independent of $\bbeta$. Using (\ref{eq:rss:2}) in (\ref{eq:rss:0}) yields
    \begin{eqnarray}
        \nonumber
        ||\by_{t} - \bX_{t} \bbeta||^2 &=& c + \tbbeta^\T \left( \bX_t^\T \bX \right) \tbbeta + \left( y_{t+1} - \bx^\T_{t+1} (\hbbeta_t + \tbbeta) \right)^2 \\
        \nonumber
        &=& c + \tbbeta^\T \left( \bX_t^\T \bX_t \right) \tbbeta + \left( z_{t+1} - \bx^\T_{t+1} \tbbeta \right)^2 \\
        \nonumber
        &=& c + \tbbeta^\T \left( \bX_t^\T \bX_t \right) \tbbeta + z_{t+1}^2 - 2 z_{t+1} \tbbeta^\T \bx_{t+1} + \tbbeta^\T \left( \bx_{t+1} \bx_{t+1}^\T \right) \tbbeta \\
        \label{eq:rss:4}
        &=& c + \tbbeta^\T \left( \bX_{t+1}^\T \bX_{t+1} \right) \tbbeta  - 2 z_{t+1} \tbbeta^\T \bx_{t+1} 
    \end{eqnarray}
    where $z_{t+1} = y_{t+1} - \bx_{t+1}^\T \hbbeta_t$ is the cross-validation error when predicting $y_{t+1}$ using a ridge solution fitted to the first $t$ observations only. Differentiating (\ref{eq:rss:4}) with respect to $\tbbeta^\T$ and solving yields
    \begin{equation}
        \tbbeta_{t+1} = z_{t+1} \left( \bX_{t+1}^\T \bX_{t+1} \right)^{-1} \bx_{t+1}.
    \end{equation}
    As $\bbeta = \hbbeta_t + \tbbeta$ it follows that the value of $\bbeta$ that minimises (\ref{eq:rss:0}) is $\hbbeta_{t+1} = \hbbeta_t + \tbbeta_{t+1}$ as claimed.
\end{proof}

\noindent We can now prove Theorem 1. \\

\noindent {\em Proof of Theorem~\ref{thm:cv:prediction:err}}.
    We will compute the cross-validation error for observation $t+1$ given $\bX_{t}$ and $\by_{t}$; as the observations are independent we may simply rearrange the order of the observations to obtain the CV error for any specific observation. We can then write the residual error for the $(t+1)$-th observation for the full least-squares fit as
    \begin{equation}
        \label{eq:error:full:ls}
        e_{t+1} = y_{t+1} - \bx_{t+1}^\T \hbbeta_{t+1}.
    \end{equation}
    Using (\ref{eq:ls:update}) from Lemma 1 in (\ref{eq:error:full:ls}) yields
    \begin{eqnarray*}
        e_{t+1} &=& y_{t+1} - \bx_{t+1}^\T \left( \hbbeta_t + z_{t+1} \left( \bX_{t+1}^\T \bX_{t+1} \right)^{-1} \bx_{t+1} \right) \\
        &=& z_{t+1} - z_{t+1} \bx_{t+1}^\T \left( \bX_{t+1}^\T \bX_{t+1} \right)^{-1} \bx_{t+1} \\
        &=& (1 - d_{t+1}) z_{t+1}.
    \end{eqnarray*}
    Solving for $z_{t+1}$ yields $z_{t+1} = e_{t+1}/(1-d_{t+1})$. Then, 
    \[
        y_{t+1} - \bx^\T_i \hbbeta_t = \frac{e_{t+1}}{1-d_{t+1}}
    \]
    and solving for $\bx^\T_i \hbbeta_t$ yields (\ref{eq:cv:prediction:err}) as claimed.
    \hfill$\Box$

\subsection{Deriving All LOOCV Models and Predictions} \label{sec-appendix-loocv}

The coefficients corresponding to each of the $i \in \{1, 2, \ldots, n\}$ separate LOOCV models (in turn, corresponding to each of the $\bX_{i}$ `left out' examples) are given by:%
\begin{eqnarray*}%
    \hat{\bm{\alpha}}_{\hat{\lambda},-i}^{(j)} &=& \hat{\bm{\alpha}}_{\hat{\lambda}}^{(j)} - \left[\frac{\bZ}{\bm{\Sigma}^{2} + \hat{\lambda}}\right]_{i} \cdot \tilde{\bm{e}}_{\hat{\lambda},i}^{(j)} \nonumber \\%
    \therefore \hat{\bm{\beta}}_{\hat{\lambda},-i}^{(j)} &=& \hat{\kappa} \, \bV \! \hat{\bm{\alpha}}_{\hat{\lambda},-i}^{(j)} \nonumber%
\end{eqnarray*}%
where $\tilde{\bm{e}}_{\hat{\lambda}}$ represents the LOOCV residuals corresponding to $\hat{\lambda}$.



\subsection{Proof of Theorem 2} \label{sec-appendix-large-p}

\noindent {\em Proof of Theorem~\ref{thm:prop:1}}. We have $\phi(\bx_i^\T \bbeta_1) = 1-\varepsilon$ for all observations with $y_i=1$ and $\phi(\bx_i^\T \bbeta_1) = \varepsilon$ for all observations with $y_i=0$. The adjusted targets are then
\begin{eqnarray*}
    z^t_i &=& \bx_i^\T \bbeta_1 + \frac{ y_i - \phi(\bx_i^\T \bbeta_1) }{ \phi(\bx_i^\T \bbeta_1) (1-\phi(\bx_i^\T \bbeta_1))}\\
    &=& {\rm sgn}(e_i) \phi^{-1}\left(\varepsilon\right) + \frac{{\rm sgn}(e_i)\varepsilon}{\varepsilon(1-\varepsilon)}\\
    &=& {\rm sgn}(e_i) \log\left( \frac{1-\varepsilon}{\varepsilon} \right)+ \frac{{\rm sgn}(e_i)}{1-\varepsilon} \\
    &=& {\rm sgn}(e_i) \left[ \log \left(\frac{1-\varepsilon}{\varepsilon} \right) + \frac{1}{1-\varepsilon} \right] \\
    &=& 2 (y_i - 1/2) \left[ \log \left(\frac{1-\varepsilon}{\varepsilon} \right) + \frac{1}{1-\varepsilon} \right],
\end{eqnarray*}
where $\phi^{-1}(\varepsilon) = \log[ (1-\varepsilon)/\varepsilon ]$ is the inverse logistic function. All the weights $v_i = \varepsilon(1-\varepsilon)$ are the same, so we have
\begin{eqnarray*}
    \bbeta^{t+1} &=& \left( \varepsilon^{-1}(1-\varepsilon)^{-1} \, \bX^\T \bW_t \bX + \varepsilon^{-1}(1-\varepsilon)^{-1} \lambda \bA \right)^{-1} (\varepsilon^{-1}(1-\varepsilon)^{-1} \bX^\T \bW_{t} \bz_t)\\
    &=& \left( \bX^\T \bX + \varepsilon^{-1}(1-\varepsilon)^{-1} \lambda \bA \right)^{-1} \bX^\T \bz_t  \\
    &=& \left( \bX^\T \bX + \varepsilon^{-1}(1-\varepsilon)^{-1} \lambda \bA \right)^{-1} \bX^\T \left( 2 (y_i - 1/2) \left[ \log \left(\frac{1-\varepsilon}{\varepsilon} \right) + \frac{1}{1-\varepsilon} \right] \right) \\
    &=& c(\varepsilon) \left( \bX^\T \bX + \varepsilon^{-1}(1-\varepsilon)^{-1} \lambda \bA \right)^{-1} \bX^\T \by \\
    &=& c(\varepsilon) \, \hat{\bbeta}^{\rm RR}_{\lambda_*},
\end{eqnarray*}
where $\lambda_* = \varepsilon^{-1} (1-\varepsilon)^{-1} \lambda$. The fourth step follows from the fact we include an (unpenalized) intercept, so the ridge regression estimates are invariant to translations of the targets. \hfill $\Box$
%
%
%

\clearpage


\subsection{Additional Results}

\begin{figure}[!h]%
    \centering%
    \includegraphics[width=\linewidth]{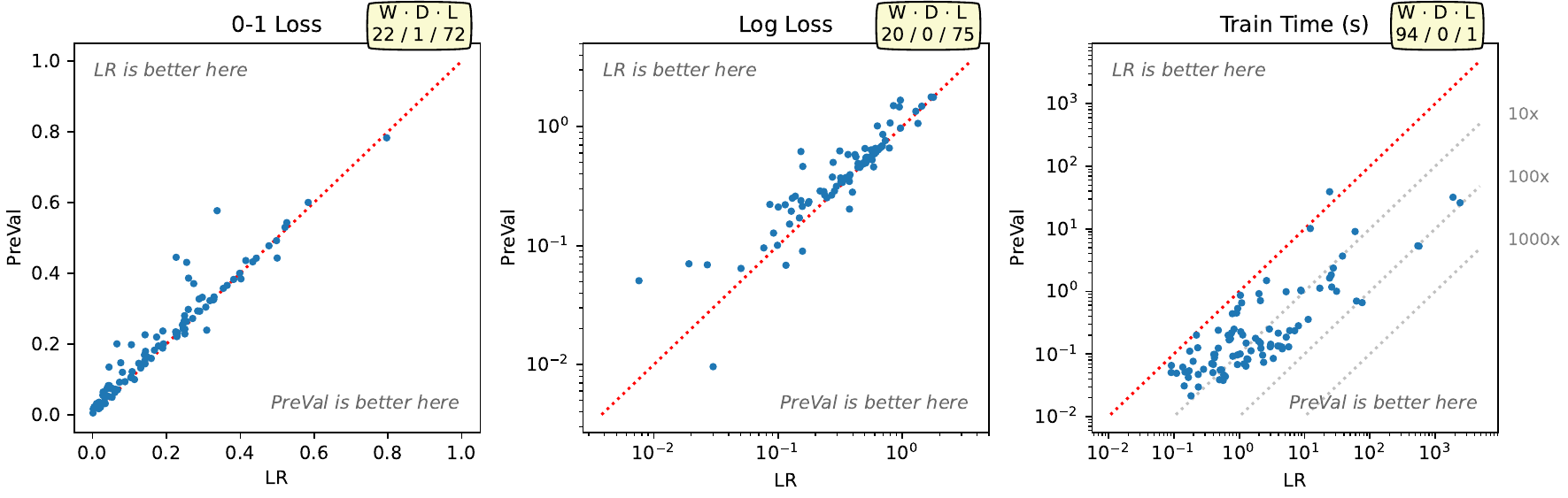}%
    \vspace{-1em}%
    \caption{Pairwise {\zo} (left), log-loss (centre), and training time (right) for {\pv} vs {\lr} on tabular datasets (original features).  On most datasets {\pv} matches {\lr} in terms of \mbox{0--\!1} and log-loss while requiring less training time.  When adding interaction terms, {\pv} more closely matches {\lr} in terms of \mbox{0--\!1} and log-loss has a greater computational advantage: see Figure \ref{fig-tabular-interactions}.}%
    \label{fig-tabular-original}%
\end{figure}

\begin{figure}[!h]%
    \centering%
    \includegraphics[width=\linewidth]{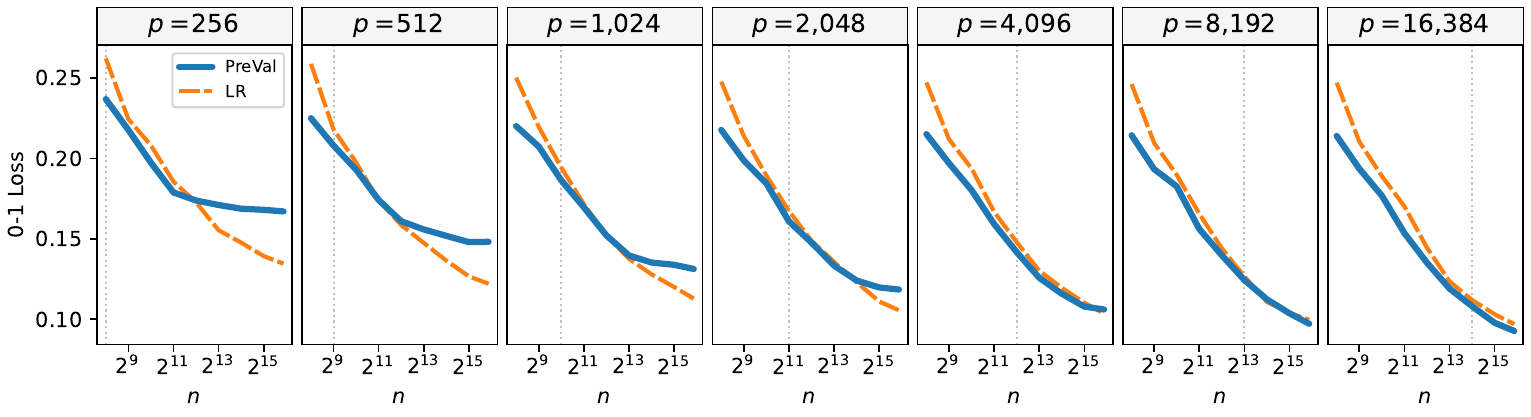}%
    \vspace{-1em}%
    \caption{Learning curves ({\zo}) for {\pv} (blue) and {\lr} (orange) for a random projection of the Fashion-MNIST dataset.  {\pv} achieves consistently lower {\zo} for small $n$.  Asymptotic {\zo} for {\pv} approaches that of {\lr} as $p$ grows.  Corresponding results for the MNIST dataset are shown in Figure \ref{fig-mnist-projection-accuracy}.}%
    \label{fig-fashion-projection-accuracy}%
\end{figure}

\begin{figure}[!h]%
    \centering%
    \includegraphics[width=0.35\linewidth]{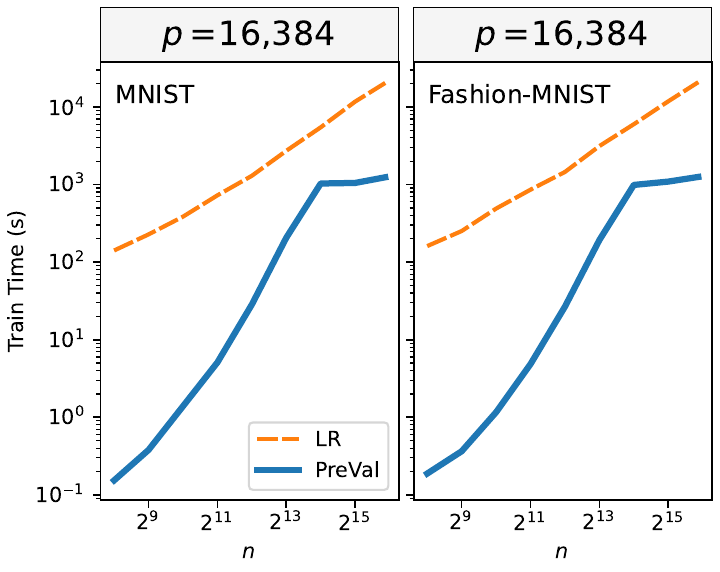}%
    \vspace{-1em}%
    \caption{Training time for {\pv} (blue) and {\lr} (orange) for a random projection of the MNIST (left) and Fashion-MNIST (right) datasets.  {\pv} is between approximately $5\times$ and $1000\times$ faster to train than {\lr}.  The shape of the curves for {\pv} reflects the fact that the underlying computational expense is proportional to $\min(n, p)$.}%
    \label{fig-fashion-mnist-projection-time}%
\end{figure}

\begin{figure}[!h]%
    \centering%
    \includegraphics[width=0.85\linewidth]{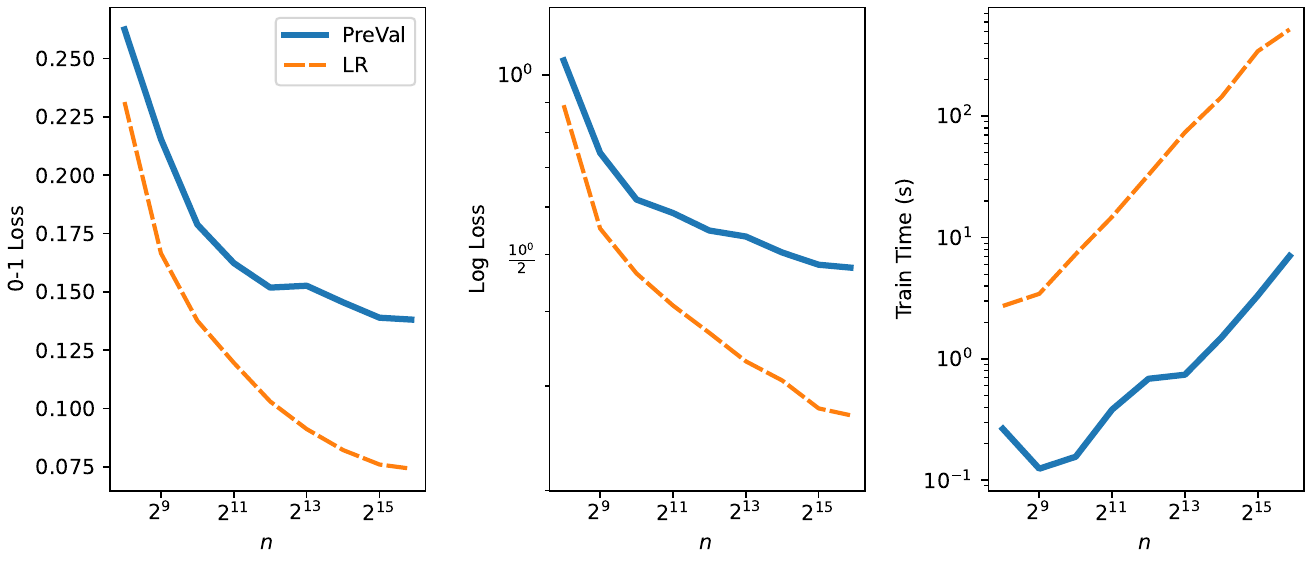}%
    \vspace{-1em}%
    \caption{Learning curves ({\zo}, left; log-loss, centre), and training time (right) for {\pv} and {\lr} on the MNIST dataset (original features).  {\lr} achieves lower \mbox{0--\!1} and log-loss, although {\pv} is more than an order of magnitude faster to train.  In contrast, {\pv} closely matches {\lr} in terms of \mbox{0--\!1} and log-loss on a higher-dimensional projection of the same dataset: see Figure \ref{fig-mnist-projection-loss}.}%
    \label{fig-mnist-all}%
\end{figure}

\begin{figure}[!h]%
    \centering%
    \includegraphics[width=0.85\linewidth]{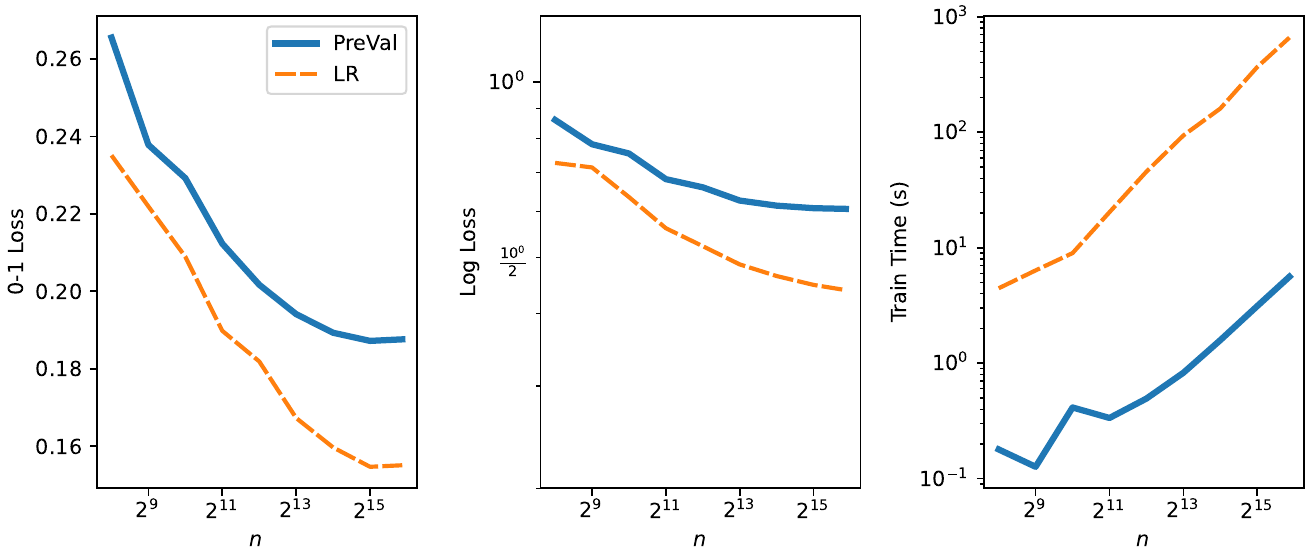}%
    \vspace{-1em}%
    \caption{Learning curves ({\zo}, left; log-loss, centre), and training time (right) for {\pv} and {\lr} on the Fashion-MNIST dataset (original features).  {\lr} achieves lower \mbox{0--\!1} and log-loss, although {\pv} is more than an order of magnitude faster to train.  In contrast, {\pv} closely matches {\lr} in terms of \mbox{0--\!1} and log-loss on a higher-dimensional projection of the same dataset: see Figure \ref{fig-fashion-projection-loss}.}%
    \label{fig-fashion-mnist-all}%
\end{figure}

\end{document}